\documentclass{article}

\usepackage[margin=1in]{geometry}
\usepackage[T1]{fontenc}
\usepackage{amsmath, amssymb, amsthm, mathtools}
\usepackage[cal=euler]{mathalpha}
\usepackage{cochineal}
\usepackage{microtype}
\usepackage{graphicx}
\usepackage[numbers,sort&compress]{natbib}
\usepackage{hyperref}
\usepackage{booktabs}
\usepackage{xcolor}

\newcommand{\akshay}[1]{}
\newcommand{\vac}[1]{}

\newtheorem{theorem}{Theorem}[section]
\newtheorem{proposition}[theorem]{Proposition}

\newtheorem{corollary}[theorem]{Corollary}

\makeatletter
\def\bvapx@title{}
\AtBeginDocument{\providecommand{\phantomsection}{}}
\let\bvapx@maketitle\maketitle
\def\maketitle{\global\let\bvapx@title\@title\bvapx@maketitle}
\newcommand{\appendixtitle}{%
  \clearpage
  \suppressfloats[t]%
  \phantomsection
  \begin{center}
    {\Large\bfseries Appendices\ifx\bvapx@title\@empty\else\space to\\ ``\bvapx@title''\fi\par}%
  \end{center}
  \vspace{1.5\baselineskip}%
}
\makeatother

\title{Cost-augmented Schr\"odinger bridges on graphs are exactly solvable:\\
a Feynman--Kac tilt replaces learned control}
\author{Akshay Balsubramani \\ {\small \texttt{akshay@vac.bio}}}
\date{}

\providecommand{\akshay}[1]{}\renewcommand{\akshay}[1]{}%
\providecommand{\vac}[1]{}\renewcommand{\vac}[1]{}%
\begin{document}
\maketitle

\begin{abstract}
The generalized Schr\"odinger bridge on a graph moves mass between two distributions while charging a cost for the states visited.
It has been approached by learning the rates of a controlled continuous-time Markov chain, with a temporal-difference penalty that restores the cost.
A state cost folds into the reference process as a Feynman--Kac tilt.
The cost-augmented bridge is then a plain bridge against the tilted reference, and the penalty is unnecessary.
The bridge is computed exactly by alternating two endpoint rescalings, each one sparse matrix-exponential application; nothing is discretized in time or learned.
The alternation converges at a rate set by the endpoint coupling alone.
For a quadratic congestion cost on time-averaged occupancies, damped best response around the exact bridge is gradient descent on a strongly convex function, and its residual bounds its error.
On a protein-folding model, a free-energy cost lowers the expected barrier of the folding paths.
On the learned approach's road network, roll-outs of the exact bridge match the target within sampling error, and on networks with millions of intersections its memory grows linearly.
\end{abstract}

\section{Introduction}
\label{sec:intro}

Transporting probability mass across a graph under operational constraints is the common form of a large class of problems: routing shipments through a supply network, assigning workers to tasks, steering a molecular simulation from one metastable basin to another.
Routing and steering ask for a time-indexed plan, which a static coupling does not supply.
For an assignment, posed on a layered graph, the plan reduces to the coupling.
In all three problems, each state visited along the way adds to the cost.
The Schr\"odinger bridge \cite{leonard2014survey} is a time-indexed plan.
Among path laws with the given endpoint distributions, it minimizes the relative entropy to a reference process.
A running cost on the visited states turns it into the \emph{generalized} bridge.

In continuous state spaces, generalized Schr\"odinger bridge matching learns a control that accounts for task-specific state costs \cite{liu2024generalizedschrodingerbridgematching}.
On a graph, recent work poses the generalized bridge as a learning problem, because it judges a direct solve prohibitive on large graphs \cite{theodoropoulos2026generalizedschrodingergraphs}.
Neural networks parameterize the potentials that define the rates of a controlled continuous-time Markov chain.
An iterative-proportional-fitting objective fits them to the endpoint marginals, but the running cost, kept outside the reference, cancels out of that objective (Section~\ref{sec:identity}).
A temporal-difference penalty, adapted from a continuous-space solver for mean-field games \cite{liu2022deep}, restores the cost.
The penalty adds a weight to be tuned, swept over six values in the supply-chain experiments of \cite{theodoropoulos2026generalizedschrodingergraphs}.
That weight trades the running cost against the terminal constraint: a larger weight lowers the peak occupancy and worsens the terminal match to the target.
The penalty's continuous-space form borrows target networks and replay buffers from deep reinforcement learning to help stabilize training \cite{liu2022deep}.

A running cost $f$ that depends only on the state folds into the reference measure as an exponential tilt.
The generalized bridge is then \emph{identically} a plain bridge against the tilted reference (Section~\ref{sec:identity}).
With the cost inside the reference nothing cancels, so no temporal-difference term is needed.
The tilt shifts the diagonal: it replaces the reference generator $Q$ by $Q - \operatorname{diag}(f)$.
The Schr\"odinger system then becomes a pair of linear ordinary differential equations, solved by matrix exponentials.
The shift leaves every edge rate in place.
The optimal control therefore reweights each edge rate by a ratio of Schr\"odinger potentials, so the controlled chain moves only along the edges of the graph.
The bridge follows by alternating two endpoint rescalings, each one sparse matrix-exponential application (Section~\ref{sec:solver}).
Both the tilt and the potential-ratio form of the control are classical; Section~\ref{sec:identity} gives their sources.
On a finite graph, the two together let sparse linear algebra solve the problem exactly.
Here exact means that nothing is discretized in time and nothing is learned: the solver applies matrix exponentials of the tilted generator to vectors numerically.
In continuous space, a bridge with a quadratic state cost is exactly solvable in a stronger sense, since its kernel has a closed form \cite{teter2025schrodinger}.

The identity does not cover costs that depend on the marginal; congestion is the canonical example.
For a congestion cost proportional to the sum of the squared time-averaged occupancies of the congestible states, we compute the congestion equilibrium by damped best response, with the exact bridge as the inner oracle.
Damped best response is gradient descent on a strongly convex function whose curvature has explicit bounds.
It therefore converges from any start when the step lies below a threshold set by the upper bound.
Near the equilibrium it also converges for larger steps, up to a threshold set by the curvature there.
Whatever the step, the residual of each iterate bounds its distance to the equilibrium (Section~\ref{sec:meanfield}).

The exact solver returns the optimal controlled rate in the form targeted by the learned approach, so it can replace that approach directly.
We evaluate it on that approach's data: a Markov state model of chignolin folding, rebuilt from the molecular-dynamics trajectories of its rare-event scenario, and the Washington DC road network of its supply-chain scenario.
On both, the exact bridge meets its endpoint marginals.
On the road network, roll-outs of its controlled rate match the target within sampling error.
On the same network, the tested dampings converge below the threshold set by the curvature at the equilibrium and fail to converge above it.
Where the largest eigenvalue of the curvature governs convergence, the residual contracts at the rate predicted from that eigenvalue.
On the assignment construction of \cite{theodoropoulos2026generalizedschrodingergraphs}, the exact solution recovers the minimum-cost assignment on every instance at concentrations within the band of that paper's main table.
The learned solver misplaces a pair at the largest size tested, where the exact solution still recovers every instance.
On chignolin, the controlled process folds in every roll-out.
A free-energy cost lowers the expected barrier of the folding paths.
The fitting iteration converges at a rate that depends on the graph only through the second singular value of the endpoint coupling.
On road networks of up to 2.07 million intersections, the solver's memory grows in proportion to the node count.

\section{The tilt identity}
\label{sec:identity}

Let $G = (\mathcal{X}, E)$ be a directed graph on a finite node set with $|\mathcal{X}| = n$, and let $Q$ be a generator supported on $E$: $Q_{xy} \ge 0$ is the rate of the transition $x \to y$ for $x \neq y$, $Q_{xy} = 0$ whenever $(x,y) \notin E$, and $\sum_y Q_{xy} = 0$.
Write $R$ for the law on $[0,\tau]$ of the chain with generator $Q$ started from a distribution of full support, let $\mu, \nu$ be distributions on $\mathcal{X}$, and denote by $q_t$ the time-$t$ marginal of a path measure $q$.
Every $q$ considered below starts at $\mu$, so the initial law of $R$ changes the objective only by a constant.
Let $f : \mathcal{X} \to \mathbb{R}$ be a running cost.

The generalized Schr\"odinger bridge is the path measure
\begin{equation}
\label{eq:gsb}
q^\star \;=\; \operatorname*{arg\,min}_{q \,:\, q_0 = \mu,\; q_\tau = \nu}
\;\; \mathbb{E}_q\!\left[\int_0^\tau f(X_t)\,dt\right] \;+\; \mathrm{KL}(q \,\|\, R)
\end{equation}

\begin{theorem}[Tilt identity]
\label{thm:tilt-identity}
Define the tilted path measure $\widetilde{R}$ by $d\widetilde{R}/dR \propto \exp\!\big(-\int_0^\tau f(X_t)\,dt\big)$. Then
\begin{equation}
\label{eq:tilt}
\mathbb{E}_q\!\left[\int_0^\tau f(X_t)\,dt\right] + \mathrm{KL}(q \,\|\, R)
\;=\; \mathrm{KL}(q \,\|\, \widetilde{R}) \;-\; \log Z,
\qquad Z = \mathbb{E}_R\!\left[e^{-\int_0^\tau f}\right]
\end{equation}
for every $q$ absolutely continuous with respect to $R$.
Consequently $q^\star$ is the plain Schr\"odinger bridge between $\mu$ and $\nu$ against $\widetilde{R}$, the normalized path measure of the semigroup $e^{tA}$ with
\begin{equation}
\label{eq:tilted-generator}
A \;=\; Q - \operatorname{diag}(f)
\end{equation}
\end{theorem}

The proof is in Appendix~\ref{app:proofs-the-tilt-identity}.

A generator has zero row sums; those of $A$ are $-f$, so $A$ is not itself a generator.
For $f \ge 0$ the semigroup $e^{tA}$ is sub-Markov: the tilted process is killed at rate $f$, so its path measure has total mass $Z$.
The tilted path measure $\widetilde{R}$ is that measure divided by $Z$.
The mass defect does not matter.
A Schr\"odinger bridge is defined against its reference only up to normalization, so the two endpoint rescalings absorb the defect exactly.
In \eqref{eq:tilt} it appears only as the $q$-independent constant $\log Z$.
The tilt is also diagonal, so it changes no off-diagonal entry and $\widetilde{R}$ is supported on the same edge set as $R$.
In particular, for $\tau > 0$, $e^{\tau A}$ has a positive $(x,y)$ entry exactly when $y$ can be reached from $x$ by transitions of positive rate, whatever the cost.
We assume that every state in the support of $\nu$ can be reached from every state in the support of $\mu$.
The block of $e^{\tau A}$ on the two supports is then positive, so the bridge exists and its endpoint scalings are positive and unique up to reciprocal constant factors \cite{menon1968matrix}.
Every instance in Section~\ref{sec:experiments} satisfies this assumption.
The optimal control inherits the edge set.

\begin{corollary}[Optimal control and support preservation]
\label{cor:control}
Let $(\varphi_t, \hat\varphi_t)$ be the Schr\"odinger potentials of the bridge against $\widetilde{R}$, so that $q^\star_t = \varphi_t \hat\varphi_t$.
The optimal controlled generator is
\begin{equation}
\label{eq:optimal-rate}
u^\star_t(x,y) \;=\; Q_{xy}\,\frac{\varphi_t(y)}{\varphi_t(x)}, \qquad x \neq y
\end{equation}
and $Q_{xy} = 0$ implies $u^\star_t(x,y) = 0$.
\end{corollary}

The proof is in Appendix~\ref{app:proofs-the-tilt-identity}.

Equation~\eqref{eq:optimal-rate} is an executable policy: a particle at $x$ consults only its out-neighbors, so the graph's topology is respected by construction, with no penalty enforcing it.
It coincides with the form targeted by the learned approach \cite{theodoropoulos2026generalizedschrodingergraphs}, so the exact solver substitutes for it without reformulating the problem.

Both the tilt and the potential-ratio form of the control are classical \cite{leonard2014survey}.
In linearly solvable Markov decision problems, whose control cost is a relative entropy, an exponential transform of the value makes the Bellman equation linear in the tilted transition matrix, with optimal transitions that reweight the uncontrolled ones by the transformed value of the next state \cite{todorov2006linearly}.
Path-integral control linearizes its Bellman equation by the same transform \cite{kappen2005}.
For a diffusion, a positive solution of a generalized Schr\"odinger system, a pair of linear equations coupled through their boundary values, yields an optimal control of the bridge with a running cost \cite{chen2021siam}.
In continuous space, iterative Markovian fitting has been extended to the generalized bridge of \cite{liu2024generalizedschrodingerbridgematching} by recasting it as a plain bridge against a Feynman--Kac transform of Brownian motion \cite{noble2026twisted}.
Discrete-time bridges on networks are taken against priors whose rows need not sum to one, with transitions of the same form \cite{chen2017robust}.
On a graph in discrete time, transport of paths regularized by relative entropy, with edge costs folded into the reference transition matrix $W$, reduces to scaling a block of its fundamental matrix $(I-W)^{-1}$ \cite{guex2019randomized}.
For random walks on directed graphs, a path cost enters as a relative entropy to the prior path law reweighted by the exponential of minus the path cost, a measure that need not be normalized \cite{eldesoukey2024schrodingers}.
Concurrent work states the tilt for controlled Markov jump processes, whose endpoint scalings are fitted by iterative proportional fitting when every jump rate is controlled \cite{srinivasan2026barsot}, and for discrete-time chains on a finite state space \cite{lollier2026quantum}.

\paragraph{The absence of a temporal-difference term.}
The learned approach holds the reference process fixed and adds the running cost to its objective as a separate term \cite{theodoropoulos2026generalizedschrodingergraphs}.
Written against the untilted reference, the potentials obey $\partial_t \varphi_t = -Q\varphi_t + f\varphi_t$ and $\partial_t \hat\varphi_t = Q^\top\hat\varphi_t - f\hat\varphi_t$, in which $f$ appears with opposite signs and cancels from $\partial_t \log q^\star_t = \partial_t \log\varphi_t + \partial_t \log\hat\varphi_t$.
The two cost terms cancel in the same way in the iterative-proportional-fitting objective of the learned approach, so on its own that objective enforces the endpoint marginals but not consistency with the cost \cite{theodoropoulos2026generalizedschrodingergraphs}.
A temporal-difference penalty restores that consistency.
It is the graph form of a loss from a continuous-space solver for mean-field games, whose forward-backward stochastic differential equations, once discretized, resemble temporal-difference learning \cite{liu2022deep}.
Written against $\widetilde{R}$, the cost is already inside $A$, so iterative proportional fitting sees it and the cancellation has nowhere to occur.
Both descriptions are of the same problem; only the second leaves the cost visible to the algorithm.

\section{An exact continuous-time solver}
\label{sec:solver}

\begin{figure}[htbp]
\centering
\includegraphics[width=\linewidth]{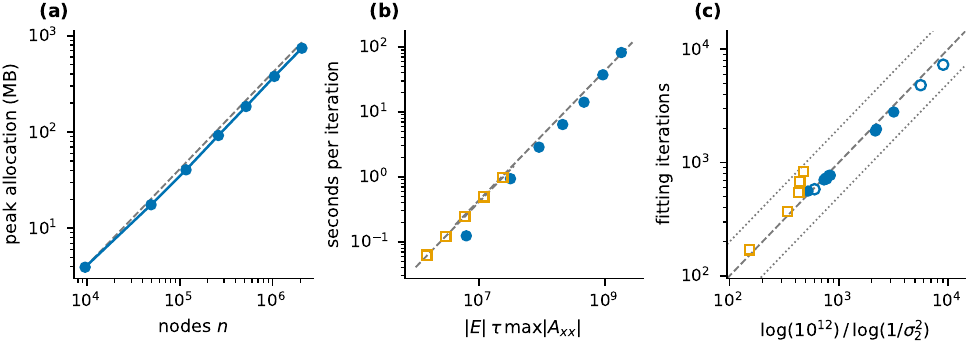}
\caption{\textbf{The number of fitting iterations follows the second singular value of the endpoint coupling, while memory and time per iteration grow with the size of the problem.} Road networks of the ROAD-FLOW benchmark from Washington DC ($n=9{,}559$) to Texas ($n=2{,}073{,}870$); 40 sources and 40 targets within 50 hops of one random center, running cost uniform on $[0, 0.5]$, horizon 100. (a) The solver's peak allocation grows in proportion to the node count, 346 to 412 bytes per node; dashed, proportional to $n$. (b) Wall-clock time per fitting iteration against the arc count times $\tau\max_x|A_{xx}|$, for the seven networks timed in one container (filled) and for Washington DC with the endpoints of \cite{theodoropoulos2026generalizedschrodingergraphs} at horizons 25 to 400, timed together in a second container (open). The dashed lines, one proportional fit per container, nearly coincide. Washington DC, the one network timed in both containers, costs $2.0\times 10^{-8}$ seconds per arc and unit of $\tau\max_x|A_{xx}|$ in the first and $4.1\times 10^{-8}$ in the second. The networks lie within a factor 2.4 of one constant (Washington DC, timed again at the end, within 1.1\% of its first timing); the horizons lie within 3\% (horizon 25, timed again at the end, within 1.3\% of its first timing). (c) Fitting iterations to a change of $10^{-12}$ in the logarithm of the scalings, against $\log(10^{12})/\log(1/\sigma_2^{2})$: the seven networks (filled), four further endpoint sets on New Hampshire (open circles) and the horizons of (b) (open squares); dashed, equality; dotted, a factor 2 either way. At the end of every solve the error contracts per iteration by $\sigma_2^2$ to within 0.03\%. The counts are 0.81 to 1.76 times the prediction and in its order on the networks (rank correlation 1.00). Every solve in (c) meets both endpoint marginals to a total variation below $3\times 10^{-13}$.}
\label{fig:scaling}
\end{figure}

Theorem~\ref{thm:tilt-identity} reduces the bridge to a plain bridge against the path measure of the semigroup $e^{tA}$, normalization aside.
The Schr\"odinger system is then linear, $\partial_t \varphi_t = -A\varphi_t$ and $\partial_t \hat\varphi_t = A^\top \hat\varphi_t$, with solutions $\varphi_s = e^{(t-s)A}\varphi_t$ and $\hat\varphi_t = e^{tA^\top}\hat\varphi_0$.
Iterative proportional fitting alternates the two endpoint conditions,
\begin{equation}
\label{eq:ipf}
\hat\varphi_0 \leftarrow \mu \oslash \varphi_0,
\qquad
\varphi_\tau \leftarrow \nu \oslash \big(e^{\tau A^\top}\hat\varphi_0\big),
\qquad
\varphi_0 \leftarrow e^{\tau A}\varphi_\tau
\end{equation}
where $\oslash$ is entrywise division.
Each half-step is one application of a matrix exponential to a vector, computed by a scaled truncated Taylor series that never forms $e^{\tau A}$ and touches $A$ only through sparse products, whose number grows at most in proportion to $\|\tau A\|_1$ \cite{almohy2011computing}.
Each application therefore takes work at most proportional to $|E|\,\|\tau A\|_1$ once that norm exceeds a small constant.
The whole time axis is handled at once, with no grid: $\varphi_t$ at any $t$ is recovered by a single further application.
Sinkhorn iterations that apply a heat kernel to vectors, without ever forming the kernel matrix, compute transport on meshes and images \cite{solomon2015convolutionald} and on graphs \cite{huguet2023geodesic}.
There the kernel is symmetric and the output is a static coupling.
The fitting iteration \eqref{eq:ipf} is the Sinkhorn iteration on the block of $e^{\tau A}$ between the two supports.
Only the entries of the scalings on the two supports enter \eqref{eq:ipf}, so when a support is small the same iteration can run on the block itself: its columns take one application each, after which every iteration is a product with a small dense matrix.
Either way the iteration converges linearly: near its fixed point the error decays by a factor $\sigma_2^2$ per iteration, where $\sigma_2$ is the second singular value of $D_\mu^{-1/2}\Pi D_\nu^{-1/2}$, $\Pi$ is the optimal endpoint coupling, and $D_\mu, D_\nu$ are the diagonal matrices of the two marginals on their supports \cite{knight2008sinkhorn,thibault2021overrelaxed}.
Started from constant scalings, the number of iterations to a tolerance $\varepsilon$ is therefore $\log(1/\varepsilon)/\log(1/\sigma_2^2)$ up to an offset set by the starting point.
On the 11 road-network problems of Section~\ref{sec:exp-scale}, the counts are 0.81 to 1.07 times this law (Figure~\ref{fig:scaling}c).
The singular value $\sigma_2$ approaches one as the coupling approaches a block-diagonal one, whose blocks do not communicate; the graph enters the rate only through $\Pi$.
Because $\Pi$ is the solution itself, $\sigma_2$ is known only once the problem is solved, so the law explains an iteration count only after the fact.
The iteration stops when the logarithm of every scaling on the support of $\mu$ changes by less than $\varepsilon$.
A change measured against the largest scaling, the rule behind the solves of Sections~\ref{sec:exp-folding} and~\ref{sec:exp-supply}, can fall below $\varepsilon$ while scalings many orders of magnitude smaller are still moving (Section~\ref{sec:exp-scale}).
Those solves read their marginal errors out of the potentials, so each reported accuracy is measured directly, whatever the stopping rule.

Across a horizon, the potential's overall scale can change by a factor of up to $e^{\|A\|\tau}$, which exceeds the range of double precision once the cost is large relative to $1/\tau$.
Propagation is therefore chunked, with the vector renormalized after each chunk.
This is exact here: both the fitting iteration and the marginal $\varphi_t\hat\varphi_t$ are invariant to a global rescaling of $\varphi$.
Evaluating \eqref{eq:optimal-rate} across a free-energy barrier likewise requires the ratio in the log domain, since $\varphi$ can span tens of orders of magnitude between basins and vanishes outside the target support as $t$ approaches $\tau$.

\section{Marginal-dependent costs}
\label{sec:meanfield}

A congestion cost depends on the marginal, so Theorem~\ref{thm:tilt-identity} does not apply: the tilt would depend on the solution.
A penalty affine in the logarithm of the number of drivers on a route depends on the marginal and still leaves a linear problem: the equilibrium of the resulting mean-field routing game solves a linearly solvable Markov decision process by one backward recursion \cite{tanaka2021linearly}.
In continuous space, a variational mean-field game with diffusion and a quadratic Hamiltonian is equivalent to minimizing, over path measures with a fixed initial marginal, a relative entropy plus a running cost of the marginals and a terminal cost \cite{benamou2019entropy}.
We charge congestion instead on the occupancy averaged over the horizon, $\bar p^{\,q} = \tau^{-1}\int_0^\tau q_t\,dt$, through the quadratic potential $F(p) = \tfrac{\beta}{2}\sum_{x \in \mathcal{C}} p(x)^2$ on a set $\mathcal{C} \subseteq \mathcal{X}$ of congestible states, for instance every state other than the endpoints.
The objective $\mathrm{KL}(q\|R) + \tau F(\bar p^{\,q})$ is strictly convex in $q$ over the convex set cut out by the endpoint constraints, since the relative entropy is strictly convex and $\tau F(\bar p^{\,q})$ is convex, $\bar p^{\,q}$ being linear in $q$.
Under the reachability assumption of Section~\ref{sec:identity} that set is not empty, so the minimizer $q^\circ$ exists and is unique.
Its linearization at a given $q$ is the running cost $\beta\bar p^{\,q}$ on $\mathcal{C}$, which depends on the state alone, so the best response to it is a bridge of Section~\ref{sec:solver}.

We reach $q^\circ$ by damped best response.
For a running cost $g$ on $\mathcal{C}$ write $\bar p(g) \in \mathbb{R}^{\mathcal{C}}$ for the time-averaged marginal of the bridge with cost $g$, restricted to $\mathcal{C}$, and $\Phi(g)$ for the optimal value of \eqref{eq:gsb} with $f = g$.
The time average is computed from the potentials by Gauss--Legendre quadrature, whose error decays faster than any power of the number of points because $t \mapsto \varphi_t\hat\varphi_t$ is entire, analytic on the whole complex plane.
From the current occupancy $p \in \mathbb{R}^{\mathcal{C}}$, the step solves the bridge with cost $\beta p$ and mixes, $p \leftarrow (1-\alpha)\,p + \alpha\,\bar p(\beta p)$.

\begin{proposition}[Damped best response is gradient descent]
\label{prop:damped-br}
Assume that every state in the support of $\nu$ can be reached from every state in the support of $\mu$.
The damped best-response step is a gradient step of size $\alpha$ on
\begin{equation}
\label{eq:h}
h(p) \;=\; \tfrac12\|p\|_2^2 \;-\; \frac{\Phi(\beta p)}{\beta\tau}
\end{equation}
which satisfies $I \preceq \nabla^2 h \preceq (1+\beta\tau/2)I$ and has the restriction $p^\circ$ of $\bar p^{\,q^\circ}$ to $\mathcal{C}$ as its unique minimizer.
For every $\alpha \in \big(0, 4/(2+\beta\tau)\big)$ the iterates satisfy $\|p_k - p^\circ\|_2 \le \rho^k\,\|p_0 - p^\circ\|_2$ with $\rho = \max\{1-\alpha,\ \alpha(1+\beta\tau/2)-1\}$.
Let $m \le L$ be the extreme eigenvalues of $\nabla^2 h(p^\circ) - I$, and for $\alpha \in \big(0, 2/(1+L)\big)$ let $\rho_\circ = \max\{1-\alpha(1+m),\ \alpha(1+L)-1\} < 1$.
For every $\epsilon \in (0, 1-\rho_\circ)$ there is a ball around $p^\circ$ that the step maps into itself, and on it $\|p_{k+1} - p^\circ\|_2 \le (\rho_\circ + \epsilon)\,\|p_k - p^\circ\|_2$.
Every $p$ satisfies $\|p - p^\circ\|_2 \le \|p - \bar p(\beta p)\|_2$.
\end{proposition}

The proof is in Appendix~\ref{app:proofs-marginal-dependent-costs}.

The last inequality of the proposition, a certificate, holds whatever the damping, so the residual reported with every equilibrium bounds its error, up to the error of the inner bridge solves.
The proof bounds $\nabla^2 h - I$ by the smaller of $\beta\tau\max_{x \in \mathcal{C}}\bar p(x)$ and $\beta\tau/2$, which is small when no state holds much of the occupancy.
The local rate $\rho_\circ$ is smallest at $\alpha = 2/(2+m+L)$; above $2/(1+L)$ the step's Jacobian at $p^\circ$ has the eigenvalue $1-\alpha(1+L) < -1$, so the equilibrium is unstable.
The global range is narrow, $0.016$ at $\beta\tau = 250$ and $0.0016$ at $\beta\tau = 2500$.
The converging dampings of Section~\ref{sec:exp-supply} exceed it by factors of 22 to 449 while staying below $2/(1+L)$, where the proposition covers them only near $p^\circ$.
For $\beta\tau$ from 10 to 250 the global range is nearly the best possible among ranges that depend on $\beta\tau$ alone.
On the path $a$--$s$--$t$--$b$ with $\mu = \delta_s$, $\nu = \delta_t$, $\tau = 1$ and $\mathcal{C} = \{a, b\}$, where $s$ and $t$ jump to each neighbor at rate $10^4$ and $a$ and $b$ return at rate $0.05$, the equilibrium occupancy splits evenly between $a$ and $b$.
For each such $\beta\tau$ its curvature $L$ comes within 1.5\% of $\beta\tau/2$, so the local threshold comes within 1.5\% of the global range.
Section~\ref{sec:exp-supply} therefore measures the convergence of its dampings from where they start, the occupancy of the plain bridge.
That section also measures $L$ on a road network and checks the threshold.

Started from the occupancy of a bridge, damped best response is the generalized conditional-gradient method applied to the primal objective, which with the step $1/(k+1)$ is fictitious play in potential mean-field games \cite{lavigne2023generalized}.
By Fenchel duality it coincides with gradient ascent on the dual, with the primal and dual of \cite{bach2015duality} exchanged; ours is the case of a quadratic regularizer.
In traffic assignment, travelers in logit stochastic user equilibrium choose routes by a logit model of their travel times.
That equilibrium solves a strictly convex program in the link travel times.
The gradient of the program is the difference between the link flows that produce those times and the flows assigned by the route-choice model at those times \cite{sheffi1985urban}.
The method of successive averages steps along minus that gradient, so it is a form of steepest descent on that program.
In Markovian route choice a route is built by choosing an arc at every node with a discrete choice model.
For that model the dual of the equilibrium problem is a smooth concave function of the link costs whose gradient is the assigned link flows minus the inverse of the link cost function \cite{baillon2008markovian,oyama2022markovian}.
The function $\beta h$ of Proposition~\ref{prop:damped-br} is the travel-time program, with $\Phi/\tau$ in the role of its expected minimum route cost and link costs $\beta p$.
Damped best response is then the method of successive averages run with a constant step.
For entropic optimal transport with a convex functional of the coupling, the continuous-time form of the same construction, a flow toward the best response computed by one Sinkhorn solve against an exponentially tilted reference, converges exponentially \cite{kazeykina2025entropic}.
In the mean-field Schr\"odinger bridge, diffusing particles interact through a nonlocal potential, which makes the problem nonconvex \cite{backhoff2020mean,eldesoukey2026generalized}.
A fixed-point iteration on its marginal flow, whose inner steps are Sinkhorn-type solves with the interaction frozen, converges locally under contraction conditions \cite{eldesoukey2026generalized}.

Linear convergence for a small enough constant step is known for generalized conditional gradient \cite{ghadimi2019conditional} and for its mean-field-game form \cite{lavigne2023generalized}.
For the method of successive averages in logit stochastic user equilibrium, a constant step below a bound set by the most negative eigenvalue of the logit mapping's Jacobian at the equilibrium converges linearly near it \cite{bagchi2026spectral}.
Near that equilibrium the fixed-point residual bounds the distance to the equilibrium path flows, up to a factor set by the Jacobian of the logit mapping \cite{bagchi2025error}.
For the bridge the curvature bound is explicit, which gives the global step range and the certificate above, valid at every $p$.
Entropic fictitious play, on an objective of the same form (a relative entropy plus a convex function of linear moments), converges linearly up to a floor that shrinks with its step \cite{nitanda2023primal}.
In entropy-regularized mean-field games with finite state and action spaces, the undamped fixed-point iteration is a contraction once the temperature exceeds a threshold set by Lipschitz constants of the game \cite{cui2021approximately}.
Here the undamped step $\alpha = 1$ lies below the local threshold $2/(1+L)$ exactly when $L < 1$.
For convex graph-structured tensor problems, generalized Sinkhorn iterations, a different algorithm that ascends the dual block by block, converge globally and, under stricter assumptions, R-linearly (with an error bounded by a geometric sequence) \cite{ringh2024graphstructured}.

\section{Experiments}
\label{sec:experiments}

We evaluate on the three scenarios of \cite{theodoropoulos2026generalizedschrodingergraphs} and use its metrics.
The rare-event and supply-chain scenarios run on that paper's data.
The first is a Markov state model of the miniprotein chignolin, a finite-state Markov chain estimated from molecular-dynamics trajectories.
We rebuild it from the trajectories \cite{iida2022molecular} used in that paper, by the construction in its appendix.
The second is the Washington DC road network of the ROAD-FLOW minimum-cost-flow benchmark \cite{kovacs2015minimumcost}, with that paper's supply and demand nodes.
Because that paper released neither its model nor a complete specification of its reference rates and simulates in discrete time, values across the two papers are comparable in kind, while comparisons within a scenario are like-for-like.
The assignment task reconstructs that paper's graph construction on cost matrices and marginals drawn here, so it is compared at matched concentration.

\subsection{Assignment}
\label{sec:exp-assignment}

For an assignment of size $K$, the construction places an intermediate node $E_{xy}$ on each of the $K^2$ pairs, with $f(E_{xy}) = C_{xy}$.
We draw each cost uniformly from the integers $1, \dots, s-1$, where $s$ is the cost scale swept below, and use uniform endpoint marginals, so the coupling read off the flux is a soft assignment.
The layered graph admits exactly two transitions.
We charge the cost once per visit to $E_{xy}$, the discrete-time form of the tilt: the argument of Theorem~\ref{thm:tilt-identity} holds with the time integral replaced by a sum, so the bridge is a two-step fitting iteration on the kernel $e^{-C}$.
The concentration of the soft assignment is measured by its row entropy, the mean entropy of the coupling's rows after each is normalized to sum to one.
Concentration is governed by the spread of $C$ relative to the entropic term, so comparing at a single cost scale would not be like-for-like; we sweep the scale and match the row entropies reported in \cite{theodoropoulos2026generalizedschrodingergraphs}.
That paper prints its assignment metrics twice and the two copies disagree at its largest size: the main table gives a row entropy of $0.17$ at $K = 20$ and the appendix table gives $0.22$, with the other five columns identical.
We match the main table's band, $0.07$ to $0.17$, and also report the appendix table's looser value.

\begin{figure}[htbp]
\centering
\includegraphics[width=\linewidth]{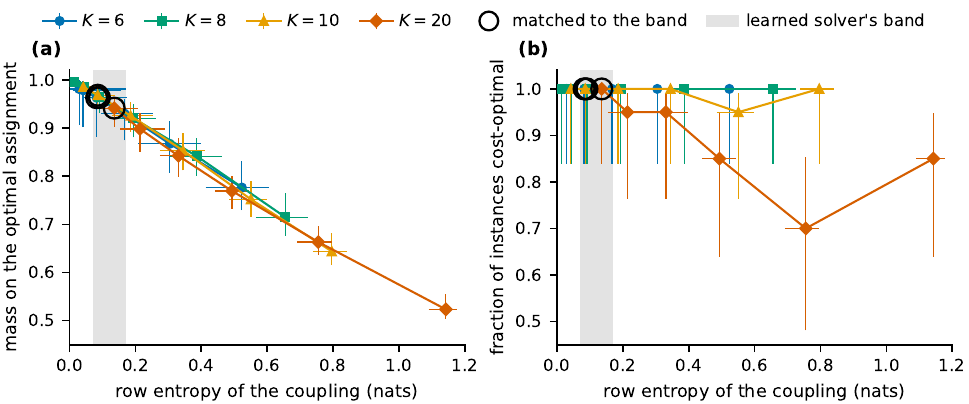}
\caption{\textbf{Inside the concentration band of the main table of \cite{theodoropoulos2026generalizedschrodingergraphs}, the exact bridge recovers the minimum-cost assignment on every instance.} Departures from the optimum appear only at higher entropy. (a) Mass the soft assignment places on the optimal assignment and (b) fraction of instances whose hard read-out is cost-optimal, against the row entropy of the coupling, for $K = 6, 8, 10, 20$. Each point averages 20 random integer cost matrices at one cost scale; along each curve the scales are 25, 50, 100, 200, 400 and 800, from right to left. The shaded band is the row-entropy range 0.07--0.17 reported for the learned solver in the main table of \cite{theodoropoulos2026generalizedschrodingergraphs}. Ringed points are the scale matched to it at each size. There the hard read-out agrees with the optimum on every row of every instance. The largest endpoint-marginal violation over the 20 instances is $1.8\times 10^{-6}$, $9.4\times 10^{-7}$, $1.9\times 10^{-6}$ and $1.0\times 10^{-6}$ for $K = 6, 8, 10, 20$, set on this layered construction by the budget of the two-step fitting iteration. These violations change no read-out. That paper's appendix repeats the table with a row entropy of 0.22 at $K=20$ in place of 0.17; at the scale matching 0.22, 19 of the 20 instances at that size are cost-optimal. Mass on the optimum falls monotonically as the entropy rises. The departures in (b) scatter at a sample size of 20 instances. Bars are 95\% intervals, bootstrap in both coordinates in (a) and Wilson in (b).}
\label{fig:assignment}
\end{figure}

Inside that band the exact solution recovers a cost-optimal assignment on every instance at every size tested (Figure~\ref{fig:assignment}).
Its hard read-out, the permutation of largest total coupling mass, agrees with the optimum on every row.
At the appendix table's $0.22$ the recovery is no longer uniform: 19 of the 20 instances at $K = 20$ are cost-optimal.
The exception is the most diffuse of the 20, with a row entropy of $0.54$ against a mean of $0.21$ over the instances at that scale.
At $K = 20$ the main table's own value, $0.17$, lies between the row entropies of the two nearest scales tested, $0.14$ with all 20 instances cost-optimal and $0.21$ with 19.
The learned solver reports 90\% agreement and a cost of 546 against an optimum of 538 at $K = 20$: the exact solution introduces no learned approximation, so the sweep isolates the error contributed by the entropic regularization alone.
At low cost scales (that is, at high entropy) even the exact bridge's hard read-out departs from the optimum,
which locates the failure in the regularized objective itself.

\subsection{Rare-event steering}
\label{sec:exp-folding}

The Markov state model is built on 500 clusters of the backbone dihedral angles of the 10-residue miniprotein chignolin, with transition probabilities estimated at a lag of 1 ns from five trajectories of 1\,$\mu$s at 300 K.
Its free energy is $F = -\log\pi$ in units of $k_{\mathrm B}T$, measured from the lowest basin state.
A state belongs to the unfolded basin when more than 80\% of its frames lie beyond 0.5 nm of C$_\alpha$ root-mean-square deviation from the folded structure, and to the folded basin when more than 90\% lie within 0.12 nm.
At the folded threshold of 0.09 nm stated in \cite{theodoropoulos2026generalizedschrodingergraphs} no state of the rebuilt model qualifies.
The threshold 0.12 nm is the smallest on a grid of 0.01 nm at which the folded basin holds at least a tenth of the frames (32\% of them, against 4\% at 0.11 nm).
At that threshold the trajectories contain five folding events and one unfolding event, so the model's folding kinetics rest on few transitions, as they must for any model built from these data.
The construction removes transitions between clusters whose mean deviations differ by more than 0.1 nm, after which the model keeps the 485 clusters of the largest connected set that remains.
The running cost is the free energy scaled by a dimensionless strength $\lambda$, so $\lambda = 0$ recovers the plain bridge.
A path's barrier is the maximum free energy along the path.
The horizon, 58 ns, is calibrated so the uncontrolled process folds 1\% of the time, against 0.55\% for the uncontrolled chain of \cite{theodoropoulos2026generalizedschrodingergraphs} over its horizon.
Bridge matching has also been used to learn Doob-transformed rates on Markov state models of other systems \cite{zhang2025scoobdoob}; on such a model the exact bridge computes those rates directly (Corollary~\ref{cor:control}).

\begin{figure}[htbp]
\centering
\includegraphics[width=\linewidth]{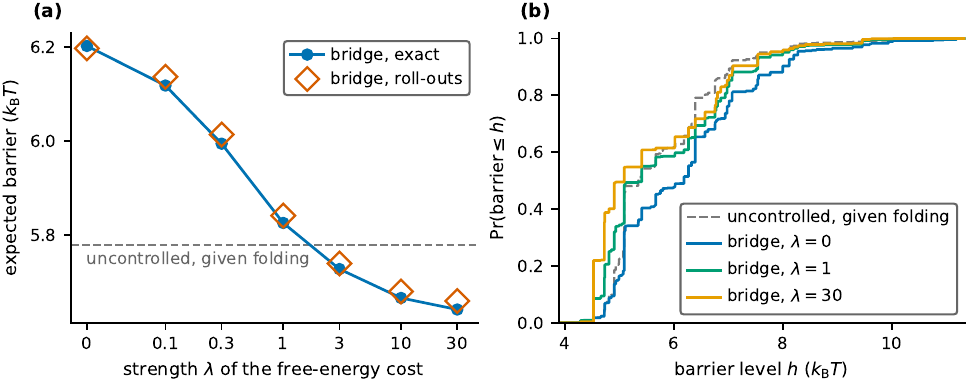}
\caption{\textbf{On a Markov state model of chignolin folding, a free-energy running cost lowers the expected barrier monotonically.} (a) Expected barrier, the maximum free energy along a path, measured from the lowest basin state, against the strength $\lambda$ of the running cost $\lambda F$ ($\lambda = 0$ is the plain bridge; the axis is linear below 0.1 and logarithmic above). Filled markers are exact, from the tilted generator restricted to each sublevel set of $F$. Open markers average over the roll-outs that fold; they lie within 0.02 of the exact values. The dashed line is the uncontrolled process conditioned on folding, exact; its roll-out estimate from its 42 folding trajectories out of 4{,}000 is 5.60. (b) The exact distribution of the barrier, $\text{Pr}(\max_t F(X_t) \le h)$.}
\label{fig:rare-event}
\end{figure}

At every $\lambda$ the bridge meets both endpoint marginals to a total variation below $10^{-16}$.
The terminal marginal of the bridge is the folded basin's equilibrium weights, so it folds in every one of 4{,}000 roll-outs at every $\lambda$, against 99.70\% and 99.36\% reported for the learned bridge without and with the cost.
The distribution of the barrier is computed exactly: the probability that a path stays in the sublevel set $\{F \le h\}$ comes from the tilted generator restricted to that set, with the bridge's endpoint scalings.
The expected barrier falls monotonically as the cost strengthens, from 6.20 at $\lambda = 0$ to 5.64 at $\lambda = 30$.
At every strength the barrier is stochastically smaller than under the plain bridge: its distribution function lies on or above the plain bridge's at every level (Figure~\ref{fig:rare-event}).
At $f = 0$ the bridge and the uncontrolled process conditioned on folding differ only through their endpoint weights (Corollary~\ref{cor:control}); here those weights raise the expected barrier from 5.78 to 6.20.
The running cost brings it back below the conditioned process at a strength between 1 and 3: the expected barrier is 5.83 at $\lambda = 1$ and 5.73 at $\lambda = 3$.
Transition path theory treats a different ensemble, the pieces of an equilibrium trajectory that leave the unfolded set and next enter the folded one \cite{metzner2009transition}.
It describes their distribution, probability current and rate through committor functions, the probabilities of reaching each set before the other.
On Markov state models of folding, the same theory gives the net flux of folding trajectories \cite{vandeneijnden2009constructing}.
On the rebuilt model no folding path has a barrier below 4.28, the free energy of the lowest unfolded state, so its barriers and the 1.80 and 1.39 reported for the learned bridge measure different models.

\subsection{Supply-chain routing with congestion}
\label{sec:exp-supply}

The network has 9{,}559 intersections and 29{,}682 arcs whose capacities follow road categories.
Reference rates are proportional to capacity and scaled to a unit mean exit rate, so time is measured in mean jump times.
The horizon is 100, the number of steps of \cite{theodoropoulos2026generalizedschrodingergraphs}.
Six supply nodes ship 6{,}360 units in total to six demand nodes.
The congestion cost falls on every node other than the supply and demand nodes, as in that paper, which penalizes the occupancy at each step where Section~\ref{sec:meanfield} penalizes its average over the horizon.

\begin{figure}[htbp]
\centering
\includegraphics[width=\linewidth]{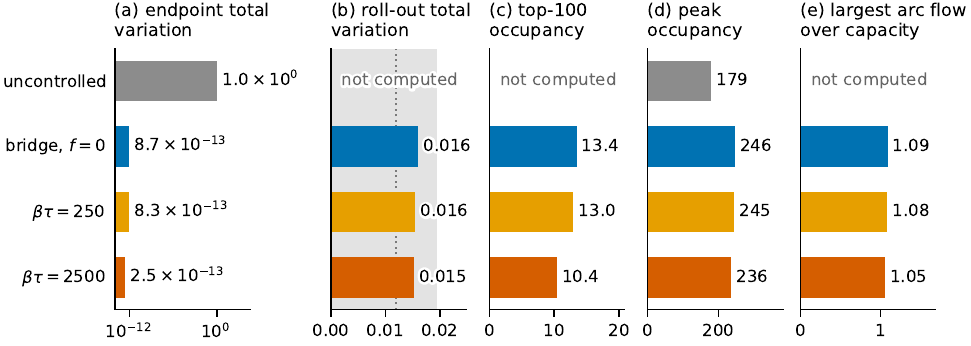}
\caption{\textbf{On the compared paper's road network the bridge meets both endpoint marginals, exactly in its read-outs and within sampling error in its roll-outs.} Washington DC roads ($n=9{,}559$ intersections, 29,682 arcs) with the six supply and six demand nodes of \cite{theodoropoulos2026generalizedschrodingergraphs}, horizon 100 mean jump times; four processes, one panel per metric, with exact values at the bar ends. (a) The larger total variation of the two exact endpoint read-outs; for the uncontrolled chain, which has no endpoint constraint, the terminal total variation. (b) The terminal total variation of 5,000 roll-outs of the controlled rate \eqref{eq:optimal-rate}. The gray band spans exact draws of 5,000 particles from the target up to their 95th percentile (0.019), and the dotted line marks their mean (0.012). (c) to (e) The compared paper's crowding metrics in expectation, per 5,000 particles: the mean time-averaged occupancy of the 100 most occupied nodes, the largest occupancy of one node at one unit time (both excluding the endpoints), and the largest flow along one arc within one unit time over its capacity. The congestion cost of Section~\ref{sec:meanfield} (bottom two bars) spreads the load. For the uncontrolled chain only the terminal total variation and the peak were computed.}
\label{fig:supply}
\end{figure}

Both endpoint marginals are met to a total variation below $10^{-12}$ (Figure~\ref{fig:supply}).
Roll-outs of the controlled rate from 5{,}000 particles end at a total variation of 0.016 from the target, within the range of exact draws of 5{,}000 particles from the target (0.012 on average, 0.019 at the 95th percentile).
On the same network and endpoints the learned solver of \cite{theodoropoulos2026generalizedschrodingergraphs} ends at 0.03 and its attraction-flow baseline at 0.27.
Its graph Schr\"odinger-bridge baseline, a dynamical formulation whose minimizer satisfies a boundary-value Hamiltonian flow on the probability simplex \cite{chow2022dynSchr}, did not complete on this network because of memory exhaustion.
Near the horizon the rate into the target grows like $1/(\tau - t)$, so a uniform time grid saturates there however fine it is.
The roll-outs therefore shrink the step geometrically over the last five time units.

Congestion spreads the load, weakly at $\beta\tau = 250$ and more at $\beta\tau = 2500$, where the mean occupancy of the 100 most occupied nodes falls from 13.4 to 10.4 particles in 5{,}000 (Figure~\ref{fig:supply}).
The largest flow along one arc within one unit of time, 1.09 times its capacity for 5{,}000 particles (1.38 for the instance's 6{,}360 units), runs into the largest demand node during the last unit of time, where the pinned terminal marginal concentrates the arrivals.
Because the congestion cost excludes the endpoints, it moves this flow only to 1.08 and 1.05.

\begin{figure}[htbp]
\centering
\includegraphics[width=\linewidth]{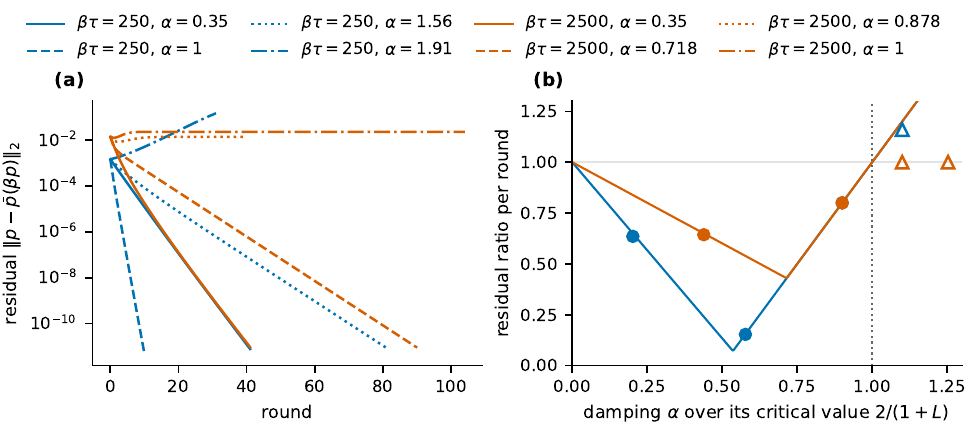}
\caption{\textbf{The tested dampings converge below the critical damping $2/(1+L)$ and fail to converge above it.} The road network of Figure~\ref{fig:supply}. (a) The residual $\|p - \bar p(\beta p)\|_2$, which bounds the distance to the equilibrium (Proposition~\ref{prop:damped-br}), against the round, at two congestion strengths. At damping 0.35 the curves for the two strengths start a decade apart and agree to within a factor 1.5 from round 10 on. (b) The ratio of successive residuals over the last three rounds of each run, against the damping in units of its critical value. Lines are the local rate $\max\{1-\alpha(1+m),\ \alpha(1+L)-1\}$ of Proposition~\ref{prop:damped-br}. Triangles mark runs that did not converge: the ratio lies above one where the residual grows and at one where it cycles. At $\beta\tau = 250$ the curvature is $L = 0.155$ (bound $\beta\tau\max_x p^\circ(x) = 1.64$), so the critical damping is 1.732; dampings 0.35, 1 and 1.56 converge and 1.91 does not. At $\beta\tau = 2500$ the curvature is $L = 1.506$ (bound $\beta\tau\max_x p^\circ(x) = 12.29$), so the critical damping is 0.798; dampings 0.35 and 0.718 converge and 0.878 and 1 do not. On the branch set by $L$ the contraction over the last three rounds is within 1.0\% of $\alpha(1+L)-1$ at both strengths. The smallest eigenvalue $m$ of $\nabla^2 h(p^\circ) - I$ is zero at both strengths, because the equilibrium occupancy is zero at the 37 congestible nodes that lie on no path from a supply node to a demand node. On the branch set by $m$, the local rate is therefore $1-\alpha$. The runs at damping 0.35 contract faster than that over their last three rounds, by 2.3\% at $\beta\tau = 250$ and 1.0\% at $\beta\tau = 2500$. Above the threshold at $\beta\tau = 250$ (damping 1.91) the residual grows by 1.16 per round, against $\alpha(1+L)-1 = 1.20$.}
\label{fig:congestion}
\end{figure}

At $\beta\tau = 250$ the largest eigenvalue of $\nabla^2 h - I$ at the equilibrium, measured by Lanczos iteration on finite differences of the best response, is $L = 0.1547$, a tenth of the bound $\beta\tau\max_x p^\circ(x) = 1.64$ from the proof of Proposition~\ref{prop:damped-br}.
The critical damping $2/(1+L) = 1.73$ exceeds one.
The plain best response converges, contracting by 0.1533 per round over its last three rounds, within 1\% of the local rate $L$; damping 0.35 converges in 42 rounds (Figure~\ref{fig:congestion}).
At $\beta\tau = 2500$, $L = 1.51$ and the critical damping falls to 0.80: the plain best response settles into a two-cycle at a residual of 0.023, while damping 0.35 converges in 42 rounds.
At 0.9 times the critical damping, an over-relaxed step of 1.56 at $\beta\tau = 250$ and a step of 0.72 at $\beta\tau = 2500$, the residual contracts by 0.800 per round at both strengths, as $\alpha(1+L)-1$ predicts.
At 1.1 times it, the residual grows by 1.16 per round at $\beta\tau = 250$, near the 1.20 the linearization gives; at $\beta\tau = 2500$ it settles into a two-cycle.

\subsection{Scale}
\label{sec:exp-scale}

The solver's memory is a few copies of the sparse generator and a constant number of length-$n$ vectors.
Each fitting iteration is two sparse matrix-exponential applications.
We measure both on the seven road networks of the ROAD-FLOW benchmark \cite{kovacs2015minimumcost}, from Washington DC (9{,}559 intersections) to Texas (2{,}073{,}870), with 40 sources and 40 targets drawn within 50 hops of one random center, a running cost uniform on $[0, 0.5]$, and horizon 100 (Figure~\ref{fig:scaling}).
The peak allocation is 346 to 412 bytes per node, 743 MB on Texas.
Timed in one container, an iteration costs $2.0$ to $4.6 \times 10^{-8}$ seconds per arc and unit of $\tau\max_x|A_{xx}|$, more on the larger networks, so over a 172-fold range of arc counts the time per iteration stays within a factor 2.4 of proportional to $|E|\,\tau\max_x|A_{xx}|$.
On Washington DC with the endpoints of \cite{theodoropoulos2026generalizedschrodingergraphs}, timed together in a second container, the time per iteration grows in proportion to the horizon from 25 to 400, to within 3\%; horizon 25, timed again at the end, comes within 1.3\% of its first timing.
The horizon needed to connect distant endpoints therefore sets the cost of a problem as much as the size of the network does.
Washington DC, the one network timed in both containers, costs $2.0 \times 10^{-8}$ seconds per arc and unit of $\tau\max_x|A_{xx}|$ in the first and $4.1 \times 10^{-8}$ in the second.

The number of iterations follows the endpoint coupling.
On 11 problems, the seven networks and four further endpoint sets on New Hampshire, the error contracts per iteration by $\sigma_2^2$ to within $5 \times 10^{-5}$.
The counts are 0.81 to 1.07 times $\log(1/\varepsilon)/\log(1/\sigma_2^2)$ and in the same order.
They range from 561 on Florida to 7{,}313 on a New Hampshire endpoint set whose coupling nearly splits in two ($\sigma_2^2 = 0.997$); Texas takes 1{,}973.
With 40 sources and 40 targets the iteration runs instead on the endpoint block (Section~\ref{sec:solver}), where each iteration is a product with a $40 \times 40$ matrix: the Texas problem takes 28 minutes on four cores, 27 of them spent on the 40 backward applications that build the block.
Where the iteration also ran on full vectors, stopped by a change measured against the largest scaling, on the networks up to Wisconsin (519{,}157 intersections), both marginals are met below $5 \times 10^{-10}$.
The terminal scalings of those solves agree to $7 \times 10^{-6}$ in the logarithm with those of a block iteration stopped by the same rule.
At $5 \times 10^4$ nodes of its own test graphs, the learned solver of \cite{theodoropoulos2026generalizedschrodingergraphs} uses 1.4 GB and about one hour; on the 49{,}109 intersections of Delaware the exact solver's peak allocation is 18 MB and its block iteration meets both marginals in 26 seconds, on different hardware.

The stopping residual matters when the endpoints lie many hops apart relative to the horizon.
At horizon 25 the endpoint block of the compared paper's endpoints spans 100 orders of magnitude.
A change measured against the largest scaling then falls below $\varepsilon$ while scalings tens of orders of magnitude smaller are still moving, leaving a terminal total variation of 0.094.
Under the logarithmic rule, as for the counts above, the iteration converges in 837 iterations, 1.76 times the asymptotic count.
Across horizons 25 to 400 the counts are 1.09 to 1.76 times the asymptotic count.

\section{Discussion}
\label{sec:discussion}

The identity of Section~\ref{sec:identity} is elementary: a change of reference measure converts a problem posed as learned stochastic control into one solved by sparse linear algebra.
The congestion cost of Section~\ref{sec:meanfield} adds an outer iteration whose residual bounds its distance to the equilibrium at every step.

The exact solver requires the reference generator explicitly.
So does the learned approach, whose algorithm takes the reference rates as input \cite{theodoropoulos2026generalizedschrodingergraphs}.
In all three scenarios considered here the reference is known: it is constructed on the assignment's layered graph, derived from a road network's capacities, or estimated from trajectories as a Markov state model, so the requirement does not bind.
Enumerating the state space is a second requirement.
A sampler for an unnormalized target on the product space $[N]^D$ faces $2^{576}$ states on a $24 \times 24$ Ising lattice.
The optimal rate there keeps the potential-ratio form of \eqref{eq:optimal-rate}; discrete Schr\"odinger-bridge samplers learn that ratio with a neural network~\cite{guo2026discrete}.

\section*{Acknowledgments}

We acknowledge the use of large language models in preparing this work.

\bibliography{exact_graph_gsb}

\appendix
\appendixtitle

\section{Proofs}
\label{app:proofs}

\subsection{Proofs for Section~\ref{sec:identity}}
\label{app:proofs-the-tilt-identity}

\begin{proof}[Proof of Theorem~\ref{thm:tilt-identity}]
The generalized-bridge objective, which is the left-hand side of \eqref{eq:tilt}, and $\mathrm{KL}(q\|\widetilde{R})$ differ by a constant independent of $q$: expanding the latter,
$\mathrm{KL}(q\|\widetilde{R}) = \mathbb{E}_q[\log q - \log R + \int_0^\tau f + \log Z]$,
which is the left-hand side plus $\log Z$.
The minimizers therefore coincide.
By the Feynman--Kac representation, $\widetilde{R}$ is the normalized path measure of $e^{tA}$: multiplying the path measure by $e^{-\int f}$ subtracts $f$ from the diagonal, leaving the off-diagonal rates unchanged.
\end{proof}

\begin{proof}[Proof of Corollary~\ref{cor:control}]
By Theorem~\ref{thm:tilt-identity} the optimizer is the plain bridge against $\widetilde{R}$, whose $h$-transform representation gives the controlled generator $A_{xy}\varphi_t(y)/\varphi_t(x)$ off the diagonal, where $A$ is the tilted operator \eqref{eq:tilted-generator}.
Since $A = Q - \operatorname{diag}(f)$ agrees with $Q$ off the diagonal,
$A_{xy} = Q_{xy}$ for $x \neq y$, which is \eqref{eq:optimal-rate}.
The support claim is immediate: $\varphi_t(x) > 0$ wherever $q^\star_t(x) > 0$, so the ratio is finite on the support of $q^\star_t$.
Moreover $u^\star_t(x,y) = 0$ whenever $Q_{xy} = 0$.
\end{proof}

\subsection{Proofs for Section~\ref{sec:meanfield}}
\label{app:proofs-marginal-dependent-costs}

\begin{proof}[Proof of Proposition~\ref{prop:damped-br}]
The bridge with cost $g$ has path law $dq_g/dR \propto \exp\{a(X_0)+b(X_\tau)-\langle g,\ell\rangle\}$, where $\ell(x) = \int_0^\tau \mathbf 1\{X_t = x\}\,dt$ is the occupation time of $x \in \mathcal{C}$.
Here $a$ and $b$ are the logarithms of its endpoint scalings on the supports of $\mu$ and $\nu$, fixed by the two marginal constraints up to adding a constant to $a$ and subtracting it from $b$.
By Lagrangian duality $\Phi(g) = \max_{a,b}\{\langle\mu,a\rangle + \langle\nu,b\rangle - \log\mathbb{E}_R\,e^{a(X_0)+b(X_\tau)-\langle g,\ell\rangle}\}$, attained at those $a, b$, so $\nabla\Phi(g) = \mathbb{E}_{q_g}[\ell] = \tau\,\bar p(g)$.
The scalings depend smoothly on $g$ because the endpoint block is positive, so $\Phi$ is twice continuously differentiable.
Hence $\nabla h(p) = p - \bar p(\beta p)$, so the step is $p - \alpha\nabla h(p)$.
Differentiating $\mathbb{E}_{q_g}[\ell]$ while $a$ and $b$ move to hold the marginals fixed gives, for $v \in \mathbb{R}^{\mathcal{C}}$,
$-v^\top\nabla^2\Phi(g)\,v = \min_{u_0,u_\tau}\operatorname{Var}_{q_g}\!\big(\langle v,\ell\rangle - u_0(X_0) - u_\tau(X_\tau)\big)$, the variance of $\langle v,\ell\rangle$ left after regressing it on additive functions of the endpoints.
This is nonnegative and at most $\operatorname{Var}_{q_g}\langle v,\ell\rangle$, with $v$ extended by zero off $\mathcal{C}$.
With $U$ uniform on $[0,\tau]$ and independent of the path, $\langle v,\ell\rangle = \tau\,\mathbb{E}[v(X_U) \mid X]$.
The law of $X_U$ is the time-averaged marginal of $q_g$, whose restriction to $\mathcal{C}$ is $\bar p(g)$.
Since $v$ vanishes off $\mathcal{C}$, the law of total variance gives $\operatorname{Var}_{q_g}\langle v,\ell\rangle \le \tau^2\operatorname{Var}\,v(X_U) = \tau^2 v^\top\big(\operatorname{diag}\bar p(g) - \bar p(g)\bar p(g)^\top\big)v$.
The last form is at most $\tau^2\max_x \bar p(g)(x)\,\|v\|_2^2$.
It is also at most $\tfrac{\tau^2}{2}\|v\|_2^2$, because $v(X_U)$ takes its values among those of $v$ and $0$, which for a unit vector lie in an interval of length at most $\sqrt 2$; a variable confined to such an interval has variance at most $1/2$.
So $\nabla^2 h(p) = I - \tfrac{\beta}{\tau}\nabla^2\Phi(\beta p)$ lies between $I$ and $(1+\beta\tau/2)I$.
Moreover $\nabla h(p^\circ) = 0$, because the minimizer of a convex objective is the best response to its own linearization.
Since $h$ is strongly convex, it has no other critical point, so $p^\circ$ is its unique minimizer and a fixed point of the step.
For any $p, p'$ the two steps differ by $M(p - p')$, where $M = \int_0^1 \big(I - \alpha\nabla^2 h(p' + s(p - p'))\big)\,ds$ is symmetric with spectrum in $[1-\alpha(1+\beta\tau/2),\,1-\alpha]$, so the step contracts by $\rho$.
With $p' = p^\circ$, $M$ tends to $I - \alpha\nabla^2 h(p^\circ)$ as $p \to p^\circ$ because $\nabla^2 h$ is continuous.
That limit is symmetric with spectrum in $[1-\alpha(1+L),\,1-\alpha(1+m)]$, so its norm is $\rho_\circ$.
Hence $\|M\| \le \rho_\circ + \epsilon < 1$ for every $p$ in a small enough ball around $p^\circ$, so the step maps the ball into itself.
Strong convexity gives $\|p-p^\circ\|_2 \le \|\nabla h(p)-\nabla h(p^\circ)\|_2 = \|\nabla h(p)\|_2$.
\end{proof}

\end{document}